# Future Trends for Human-AI Collaboration: A Comprehensive Taxonomy of AI/AGI Using Multiple Intelligences and Learning Styles

**Andrzej Cichocki and  Alexander P. Kuleshov**

**Skolkovo Institute of Science and Technology (SKOLTECH), Moscow, Russia**

**Abstract**

This article discusses some trends and concepts in developing new generation of future Artificial General Intelligence (AGI) systems which relate to complex facets and different types of human intelligence, especially social, emotional, attentional and ethical intelligence.  We describe various aspects of multiple human intelligences and learning styles, which may impact on a variety of AI problem domains. Using the concept of "multiple intelligences" rather than a single type of intelligence, we categorize and provide working definitions of various AGI depending on their cognitive skills or capacities. Future AI systems will be able not only to communicate with human users and each other, but also to efficiently exchange knowledge and wisdom with abilities of cooperation, collaboration and even co-creating something new and valuable and have meta-learning capacities.  Multi-agent systems such as these can be used to solve problems that would be difficult to solve by any individual intelligent agent.

*Key words: Artificial General Intelligence (AGI), multiple intelligences, learning styles, physical intelligence, emotional intelligence, social intelligence, attentional intelligence, moral-ethical intelligence, responsible decision making, creative-innovative intelligence, cognitive functions, meta-learning of AI systems.*

## 1. Introduction

Both human intelligence, as defined by innate, biological intelligence, as well as artificial intelligence (AI), commonly defined as machine intelligence, have been hot topics in a wide spectrum of scientific literature [see e.g., 1-13, 50]. In this paper, we explore how multiple aspects of human intelligence and various learning styles may further inspire or promote the development of new kinds of improved intelligent multi-agent systems. We also seek to provide an introduction on how different AI-systems can be categorized and ranked depending on their cognitive skills and learning styles.

Progress and even breakthroughs in AI have typically demonstrated that AI systems have the ability to perform (often very well) at solving specific problems or tasks such as recognition, classification, ranking, prediction, clustering, segmentation, playing games, like Go/Jeopardy, and even creation of artwork, like music or paintings [see e.g., 5,14,15, 52-55]. However, as AI systems evolve and expand, it is increasingly clear that their full potential lies beyond merely solving well-defined problems or performing tasks within preset parameters. Rather, many AI systems of the future will interact with other AI sub-systems (like smart robots or multi-agents) alongside human users to solve dynamically changing and complex problems. For this type of interaction to take place, multi-agent systems must have the ability to continuously learn, review, and evolve their interaction strategies during ongoing communication. In other words, research in AGI may go far beyond a single mental-intellectual or logical-mathematical intelligence, towards the concept of multiple intelligences. We refer to this phenomenon as "***AGI with multiple intelligences"*** as until now most AI methods and approaches are based on one single type of intelligence and perform only one specific task or a set of a few closely related tasks, without fully exploring and implementing sophisticated cognitive skills, emotional-social intelligence (ESI), and responsible group decision making.

In this paper, we mostly consider AI systems which are, in general, in a form of a Multi-Agent System (MAS), e.g., some kind of smart humanoid robots or computerized systems composed of multiple interacting agents (see Figure 1). This is closely related to the concept of Distributed Artificial Intelligence (DAI), which is an important subfield of AI research, dedicated to the development of distributed solutions for specific problems [see e.g., 14, 15, 21, and 22].

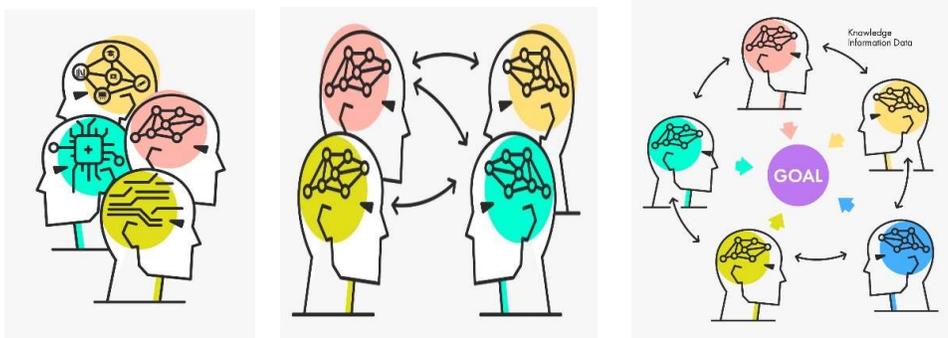

Figure 1  Multi-agents in future AGI could not only communicate and exchange data and information, but may have also ability to coordinate, cooperate and/or collaborate and exchange knowledge or even wisdom. (Note that *wisdom* can be considered as a specialized and applied *knowledge* that acts to filter/select the *knowledge* that *is* best used to extract the appropriate *information* from data. *In other words, wisdom is* the ability to put that *knowledge* to good use or the ability to determine which *facts* are relevant in certain situations.).

## 2. AI as Multidisciplinary Research

AI is grounded in, advanced by research from multiple disciplines, inspired, and driven by Human Cognitive Science, Systems Neuroscience and the Computational Sciences (see Figure 2).

Particularly, human cognitive science and systems neuroscience play key roles in the development of new AI concepts and smart/intelligent systems. In fact, AI research spans the intersection of many fields including human brain science, computer science and applied mathematics [5-7, 58]. That being said, human cognitive science is a highly interdisciplinary area in itself, exploring ideas and methods from biology, psychology, philosophy, and linguistics. Fundamental human cognitive processes are related to higher-level functions of the human brain and encompass language, imagination, perception, and planning [7, 14, 19, 41, 57, 58].

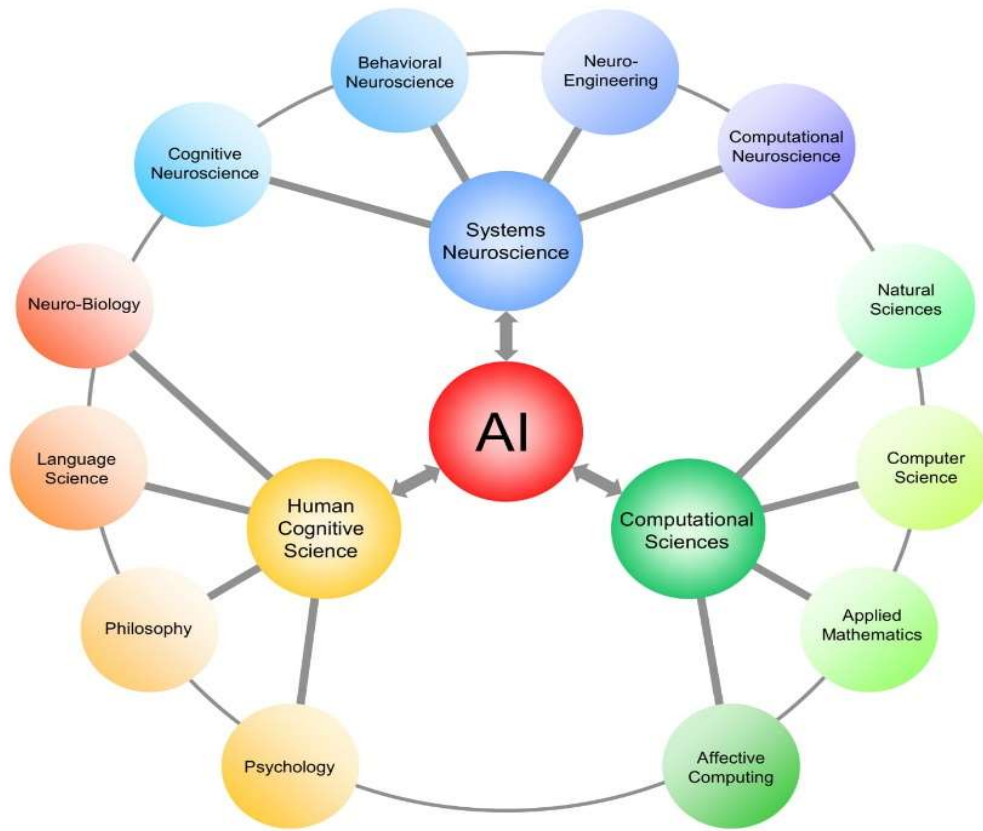

Figure 2 AI as multidisciplinary research of three areas: Cognitive Science, Systems Neuroscience, and the Computational Sciences. On the other hand, AI methods allow us to better understand systems neuroscience and human cognitive science and even develop new computational and mathematical tools.

Cognitive skills (functions, abilities or capacities) allow us to receive, select, store, transform, develop and retrieve information and knowledge that we have received from external stimuli. A better understanding of such cognitive functions and processes in the human brain would allow us to implement them in future generations of AGI systems more effectively and extend their flexibilities and applications (see Section 9).

## 3. AI Subdomains: Current Key Applications

In recent years, AI research has made tremendous progress which has already found applications in many fields, from computer vision (CV) (e.g., machine vision, robot vision) and pattern recognition (classification, clustering), to areas like robotics and intelligent agents (R/IA), Natural Language Processing (NLP) (*natural language generation (NLG), natural language understanding (NLU), machine translation, sentiment analysis, information retrieval and extraction*), Speech Recognition and Synthesis (SR/S), Planning, Scheduling and Optimization (PSO), Knowledge Representation and Reasoning (KRR) and Expert Systems (ES) (see Figure 3).

Currently, Machine Learning (ML) and its subdomains – Deep Learning (DL) and Artificial Neural Networks (ANN) – play a key role in AI research. Many researchers consider ML to be an important subdomain of AI. However, in our opinion, not all ML algorithms and methods, and not even all ANN, can be classified as part of AI since not all methods in machine learning mimic human intelligence. The same is true for all other domains, like expert systems, NLP or data mining (Figure 3).

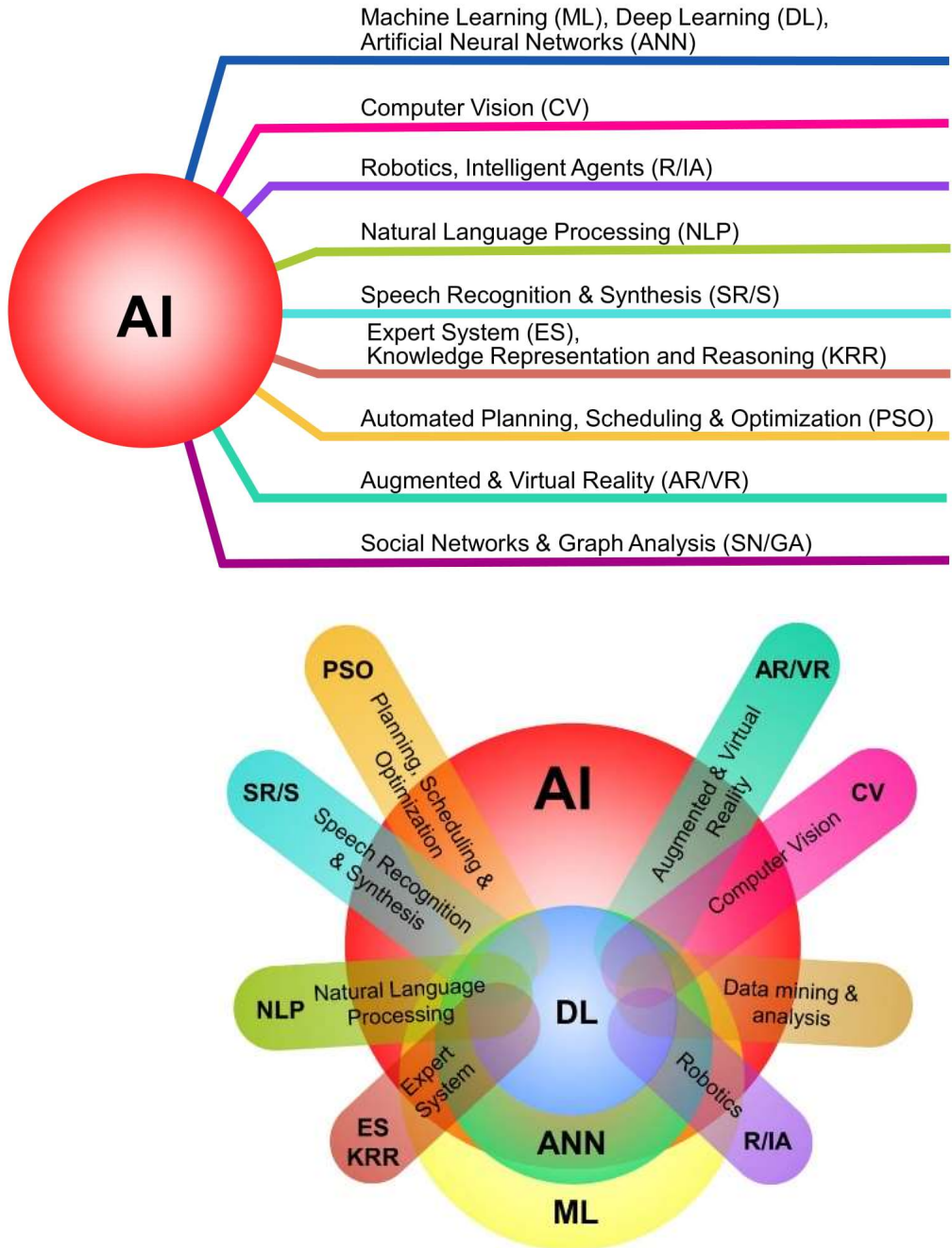

Figure 3 Core areas related to AI: AI can be considered as a wide collection of different technologies, rather one independent field. Not all methods used in machine learning (ML) belong to AI.

## 4. Multiple Intelligences

In Figure 4, we illustrate seven vital multiple human intelligences: Physical intelligence (physical quotient – PQ), mental-intellectual intelligence (verbal-logical-mathematical, – IQ), emotional and social intelligence (referred to as EQ and SQ), creative Intelligence (CQ), innovative intelligence (INQ) and moral and ethical intelligence (MQ) (cf. also [33, 38-43, 51]).

- *Physical Intelligence or Physical Quotient PQ*, also known as bodily-kinesthetic intelligence, is an intelligence derived or learned through physical, tactile and practical learning such as sports, dance or craftsmanship. Physical *intelligence* is an important aspect of personal effectiveness and physical performance.

- *Mental/intellectual Intelligence aka Intelligence Quotient – IQ* is the mental ability involved in language, mathematical analytical skills, logical reasoning, perceiving relationships and analogies, calculating, data interpretation, verbal abilities, visual and spatial reasoning, classification, and pattern detection and recognition.

- **Emotional intelligence (Emotional Quotient – EQ)** is the ability to perceive, assess, generate, understand and control emotions. EQ also involves the regulation of emotions to promote further emotional and intellectual growth. The concept of EQ was conceptualized and investigated by Michael Beldoch and later popularized by Daniel Goleman, among others [35, 36].

- *Social Intelligence (Social Quotient – SQ),* is the capacity to understand other humans and to act both rationally and emotionally in relation with others. SQ is important particularly when forming social bonds and when working within a team. This type of ability does not necessarily need to be limited to humans, but could also potentially describe the intelligence of a network of intelligent multi-agents, which need to perform complex tasks or solve specific problems jointly, while resolving the various conflicts, which may arise from working within a group.

- *Creative Intelligence (CQ)* is the capability to create or to act of conceiving something original or unusual, while *Innovative Intelligence (INQ)* is the implementation of something that has never been made before and is recognized as the product of some unique insight. Note that creativity *is characterized by generating something new, either a new idea, concept, process or method, while* innovation *employs* creativity *to enhance performance or feature of a specific product, process, person, team or organization.* Creative and innovation intelligence can be integrated as **CINQ** quotient, since *innovation* goes hand in hand with *creativity* and there is no *innovation* without *creativity*. The CINQ can be considered as higher form of human intelligences because they go beyond knowledge recall and extends into knowledge creation.

- *Moral and Ethical Intelligence* (aka **MQ**) is defined according to *Lennick and Kiel* [33] as "*the mental capacity to determine how universal human principles should be applied to our personal values, goals, and actions*". Usually, being morally and ethically intelligent means not only just assessing what is right and what is wrong, but also having the courage to do what is right and prevent both oneself and others from doing the wrong things [32-34].

*As regards Social Intelligence – SQ*, in particular, this was first conceptualized by psychologist Edward Thorndike (see e.g., [37]), but was later reinvented, extended and popularized by many psychologists, especially Howard Gardner [8, 9] and Daniel Goleman [36]. Gardner proposed and investigated eight human multiple intelligences, out, of which the two most important ones are: *intrapersonal intelligence* and *interpersonal intelligence*, which correspond to EQ and SQ intelligences in the above schema (see Figures 4 and 6).

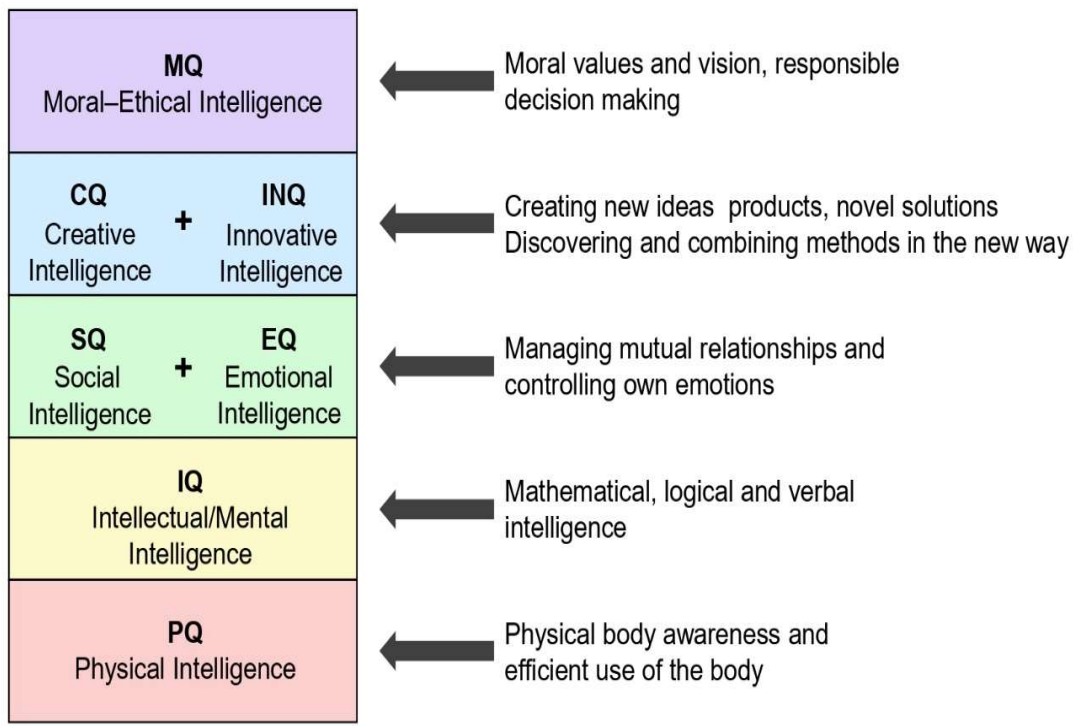

Figure 4  Seven human multiple intelligences, that are, in general, strongly correlated and mutually dependent.

According to the theory of multiple intelligences proposed by Gardner, at least eight different types of intelligence exist: Logical-mathematical (reasoning, number smart), Visual-spatial (picture smart), Verbal-linguistic (word smart), Musical-rhythmic and harmonic (sound smart), Bodily-kinesthetic (body smart), Naturalistic (nature smart), Intrapersonal (self-smart), and Interpersonal (social smart) [8, 9]. Most humans have all of these intelligences, but not all of them are developed in all of us equally or sufficiently well, therefore we often do not use them effectively. A person with only one or two intelligences, which are well developed, may have difficulties to function in the world: such is the case, for example, of many people with Autism Spectrum Disorder (ASD).

Howard Gardner defined intelligence as "*the ability to find and solve problems and create products of value in one's own culture*". Pei Wang [3, 4] defined intelligence as "the capacity of an information-processing system to adapt to its environment while operating with insufficient knowledge and resources."

***Gardner's theory of multiple intelligences*** has come under some criticism from researchers in education, psychology and philosophy [10, 39-42]. These critics argue that Gardner's definitions of multiple intelligences are too broad and mostly represent what could be called talents, abilities, preferences or personality traits. Others, meanwhile, argue that his definition was not broad enough, as he did not include ***spiritual intelligence*** in his list (encompassing concepts such as love, generosity, openness, courage, self-discipline, forgiveness, compassion, detachment, sense of purpose). However, this was mainly due to the challenges of codifying quantifiable scientific criteria for this type of intelligence, which Gardner rigorously investigated. We do, however, note that spiritual intelligence is considered by some researchers as the most sophisticated form of human intelligence since it is related to the formation of higher meanings and human values. Even if some of multiple intelligences could be controversial or do not exist for humans [1, 7], we believe that they are very useful not only to categorize but also to develop new AI systems.

Why, then, is multiple intelligence theory so interesting and important in AI research and development? First of all, it allows AI systems to learn a variety of different tasks and solve different, or even unrelated sub-problems at once. Moreover, possessing multiple intelligences can draw multi-agents back into a specific learning style, which may be most appropriate to the task at hand. Furthermore, through using their different intelligences together, multi-agents can direct their attention to more specific tasks and problems and this may help to increase their efficiency of learning, and consequently their performance and/or better decision making [23-26].

AI systems can be classified according to their functionality and ability (see for details Figure 5 (a) and (b)) [4, 44-51].

(a)

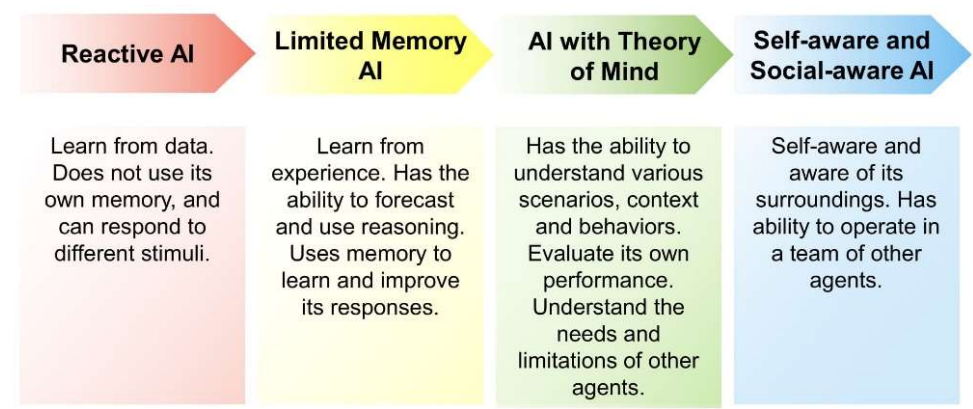

(b)

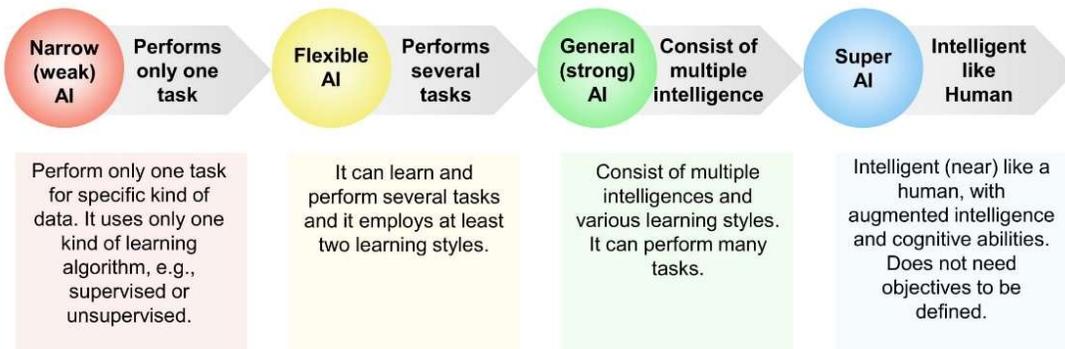

Figure 5 Types (classification) of AI and future trends in development of general AI: (a) based on functionality, (b) based on capability. Note that super AI (that may surpasses human intelligence and ability) is purely speculative at this point, i.e., it is not likely to exist for an exceedingly long time (if at all [51]).

## 5. Learning Styles and Machine Learning (ML) Algorithms

The main difference between *multiple intelligences* and *perceptual learning styles* is that multiple intelligences represents different intellectual and cognitive *abilities*, while corresponding perceptual (sensory) learning styles are the different *ways* in which a human or an intelligent agent approaches and learns a specific ability to solve problems or execute desired tasks depending on available sensory data. The Sensory Learning Style, also known as the VAK, uses the three main sensory receivers: Visual, Auditory, and Kinesthetic (see Figure 6 (a)).

Multiple intelligences can be learned – or at least improved and enhanced – via systematic and continuous learning using suitable sensory/training data and appropriate social interactions. It should be noted that certain learning styles can help to build social skills in multi-agents, that is to learn and to develop some knowledge and experience who and what is around them and how properly communicate and interact socially in order to do some tasks/actions or to make responsible decisions (see Figure 6 (b)).

(a)

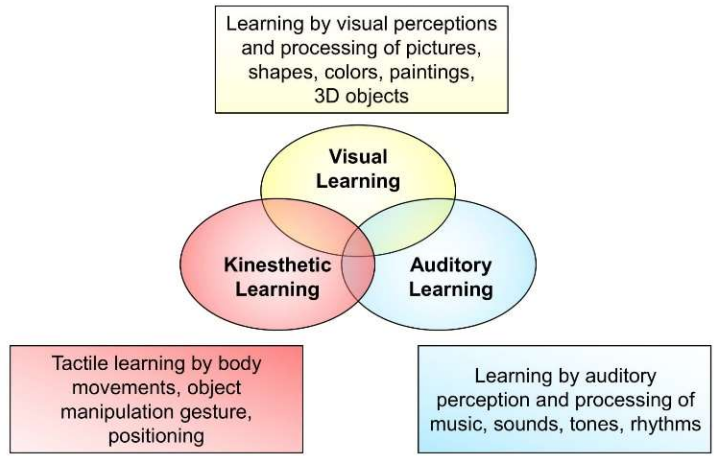

(b)

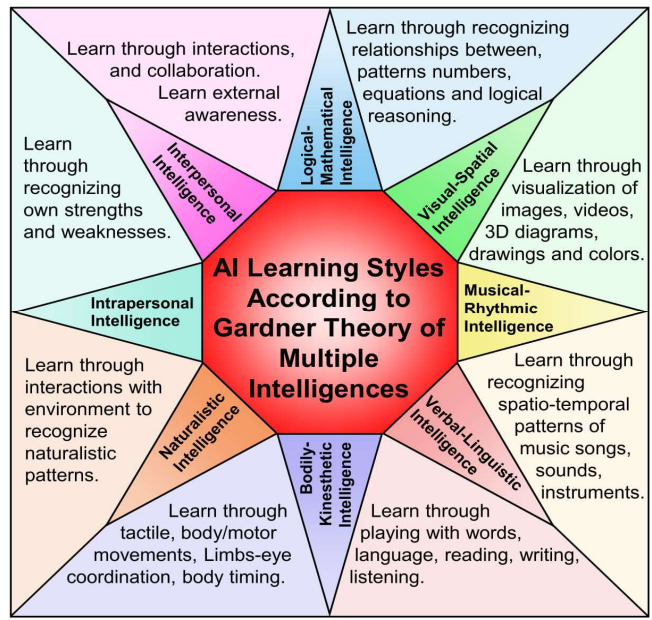

Figure 6 (a) Three main human perceptual (sensory) learning styles and (b) eight different learning styles and their corresponding multiple intelligences, as formulated by Gardner.

Moreover, by learning from the different modality of data, we can improve considerably the performance. For example, by the integration of audio data with visual data (lips movements) speech recognition can be dramatically improved in a noisy environment (mechanism that called neural binding in the neuroscientific literature [56]).

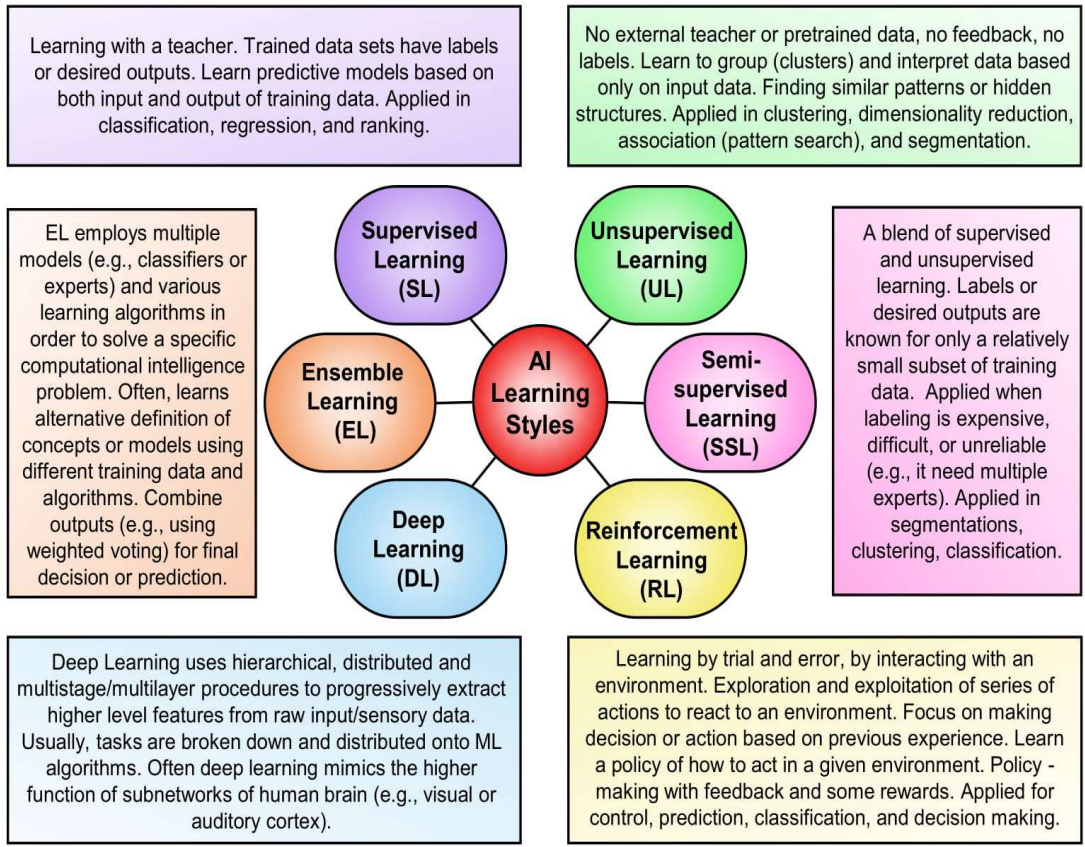

Figure 7 Basic description and features of six fundamental learning approaches used in ML and AI systems (cf., e.g. [52]).

Some of the multiple intelligences have already been explored in commercial AI systems. For example, the BAIDU AI Composer is now used to compose creative music inspired by artistic paintings. Just to mention a few others, the AIVA (Artificial Intelligence Virtual Artist) system has a musical intelligence with the ability to compose original music for films, the Intelligent Atlas robot developed by Boston Dynamics possesses impressive bodily-kinesthetic (physical) intelligence, and DeepMind's AlphaGo, which employs a Monte Carlo tree search combined with reinforcement learning algorithm, possesses sophisticated logical-mathematical intelligence to play, almost perfectly, a complex game Go. However, as of now, none of these AI systems can perform two or more quite different cognitive tasks.

In current AI systems, we extensively use six basic ways of learning: Supervised learning, unsupervised learning, semi-supervised learning, reinforcement learning, ensemble learning and deep learning (see for detail Figure 7). Particularly, important and useful for our concepts and models are: Ensemble learning, deep learning algorithms, and deep reinforcement learning [5, 6, 15, 20, 21] (see also Figures 16-20 in Appendix for more details).

## 6. AI Systems with Multiple Intelligences

We now provide a new categorization and working definitions of AI systems (multi-agents) depending on their abilities, flexibility and level/type of intelligence as follows (see also Figure 8).

***AI with Physical Intelligence abilities (AI-PQ)*** is an AI system implemented not only in software but also physically in hardware (e.g., as an electronic neuromorphic chip), which can perform specific tasks on-line or in near real-time with the ability to demonstrate good physical efficiency, that is: low power consumption, high speed, low latency, robustness and resilience to changing conditions and environmental conditions (like temperature, pressure or humidity). Such an AI system should have also the ability to control and automatically optimize power consumption depending on tasks and preferences.

***AI with mental or intellectual abilities (AI-IQ)*** **is** a computerized AI system, which can perform some logical, mathematical, analytical, and/or verbal tasks with the abilities of analytical skills, logical reasoning, pattern recognition (the ability to relate or recognize multiple patterns or events) and/or the ability to store and retrieve information.

***AI with Emotional Intelligence abilities (AI-EQ)*** is an AI system, which possesses self-awareness, self-assessment, *self-regulation (or self-management)*. In other words, AI-EQ has the capacity to evaluate/assess its own performance. It also should have reliability and robustness of performance for specific tasks, for example robustness in respect to noisy, corrupted and incomplete data sets (i.e., efficient treatment/ processing of incomplete data). The AI-EQ should also have the ability to self-assessment of own performance depending on the noise level or incompleteness of sensory data sets.

**Remark 1**: *It should be noted that our **AI-EQ** should not be confused with **Emotional AI**. Emotional AI systems refer to technologies that use affective computing and AI methods to sense, detect and classify human emotions and behaviors. Affective computing, in general, is the study and development of AI systems and devices that can recognize, interpret, process, and simulate human affects. However, affective computing aims mostly to enable AI systems to understand the emotional states expressed by human subjects* (see e.g., [30, 31]).

*It should be noted that in this paper, we consider a more general scenario, where AI-EQ is defined as an AI system, which possesses its own self-awareness, self-assessment and self-management (self-regulation).*

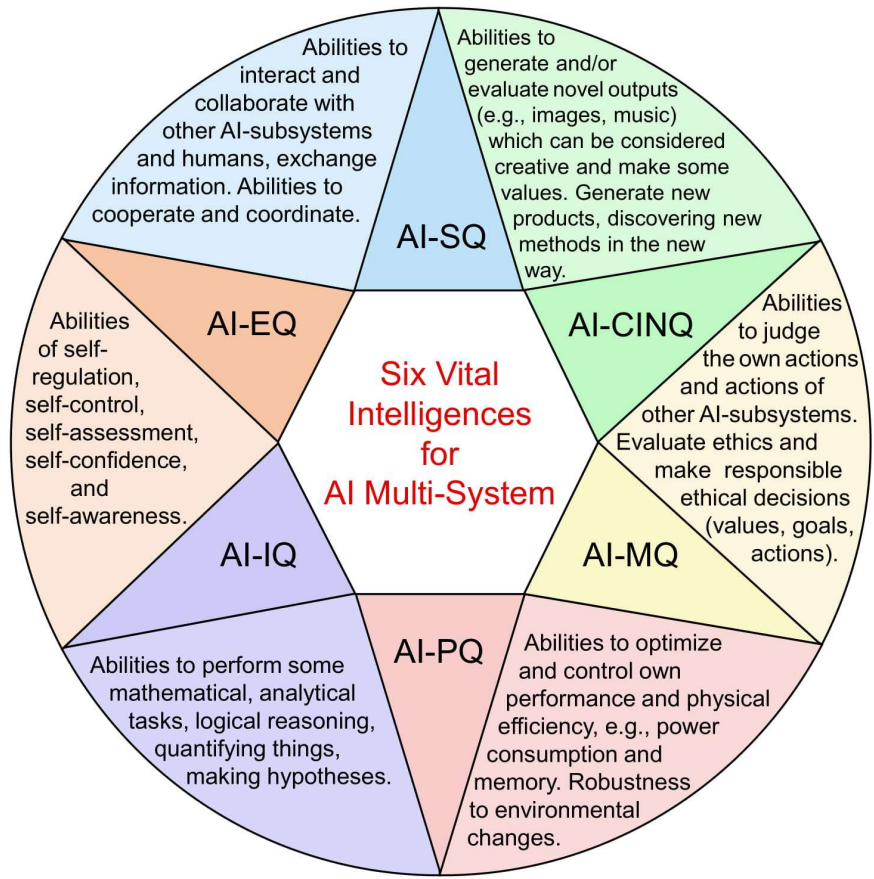

Figure 8 Taxonomy of AI (multi-agents) with the six intelligences: Physical intelligence (PQ), Intellectual intelligence (IQ), Emotional intelligence (EQ), Social intelligence (SQ), Creative + Intellectual intelligence (CINQ) and Moral-Ethical intelligence (MQ).

***AI with Social Intelligence abilities (AI-SQ)*** is an AI system, which has the ability to interact and communicate with human and/or other AI sub-systems (e.g., Deep Neural Networks (DNNs), intelligent robots, multi-agents) and exchange the information and knowledge and support each other. Moreover, such AI-SQ has the ability to coordinate, cooperate and even collaborate with other AI sub-systems (intelligent agents). For example, for the ensemble of DNNs which have the ability to not only communicate but also cooperate and/or collaborate to perform joint complex tasks in an optimized way.

***AI with Computational Creativity and Innovation (AI-CINQ)*** is an AI system that has the capacity to both generate, implement and evaluate novel product or outputs, e.g., images, music or videos, which would, if produced by a human, be considered creative and to have value and purpose, or conform to common sense (see also Figures 9).

(a)

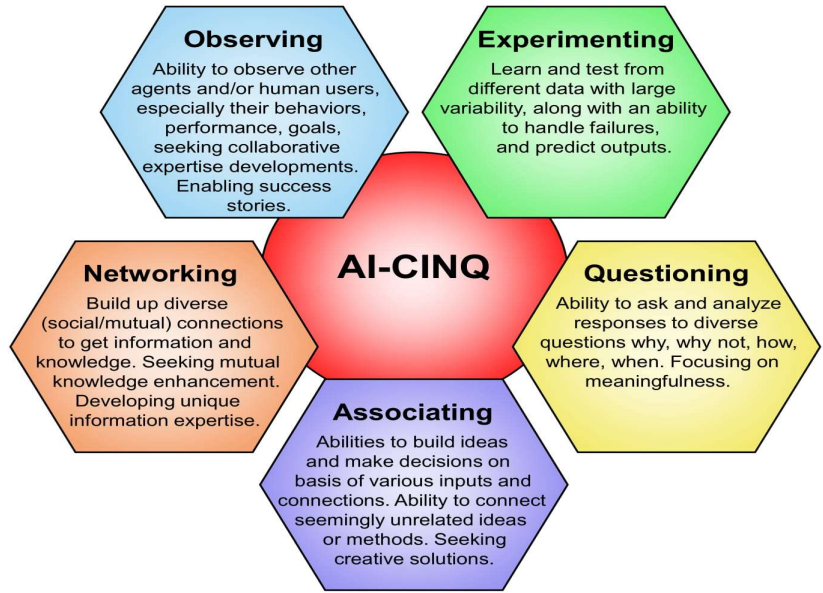

(b)

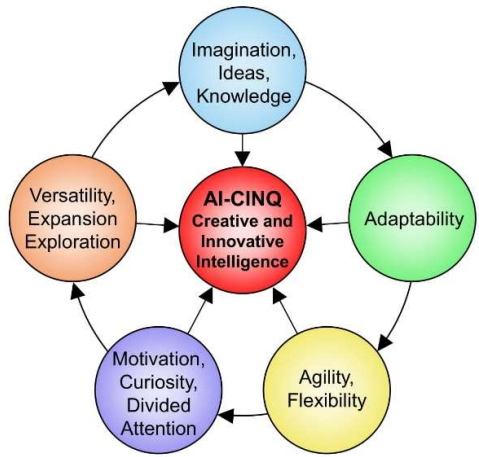

(c)

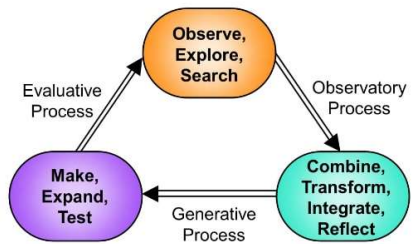

Figure 9 (a) and (b) components of creative and intellectual intelligence (**AI-CINQ**), and (c) creative-innovative intelligence process.

**Creative intelligence (AI-CQ)** involves the generation of novel and useful ideas, while **innovative intelligence** (**AI-INQ**) is concerned with the work required to make these ideas valuable or in other words entails the implementation of these ideas into new products or processes.

Summarizing, **AI-CINQ** (AI with **C**reativity and **IN**novation Quotient **CINQ**) is defined as an AI system which has capacity of solving problems and/or generate new products, processes or outputs by discovering and combining ideas and methods in the new way.

**Remark 2:** *Creative and innovative solutions or ideas can be produced in several ways: (1) Novel (non-trivial, unexpected) combinations of familiar ideas; (2) Nonlinear or multilinear transformation of original data sets into higher dimensional spaces, so that new structures can be generated, which could not have arisen before; (3) Generation of novel ideas by the exploration of structured conceptual spaces. Note that the computational creativity (also known as artificial creativity, creative computing or creative computation) is a closely related multidisciplinary endeavor that can be considered as the intersection of the fields of artificial intelligence, cognitive psychology, philosophy, and the arts* [43-46, 57, 58].

*AI with Ethical and Moral Intelligence (AI-MQ)* is defined as an AI system which has the ability not only to judge its own actions and actions of others (whether agents or humans) from the point of view of ethics, but also have executive power to make responsible decisions to prevent "wrong" doing. In other words, AI-MQ should have some kind of self-awareness and executive power not only to judge or asses what is "right" and "wrong" but also should have ability to take action to *do* what is right. *AI-MQ intelligence* would be most challenging to implement from of all *intelligences which we discuss in this paper* though, however, the most valuable for humanity.

**Remark 3**: *While ethics and morals both relate to "right" and "wrong" behaviors and are therefore often used interchangeably, we do differentiate between them in a substantial way: While* **morality** *is something normative but usually personal,* **ethics** *is the standard of "right and wrong" which is established by a certain community, culture, or social setting (e.g., codes of conduct in workplaces). In other words, ethics refers to rules provided by an external source, whereas morals refer to an individual's own principles regarding what is "right" and "wrong"* [32-34].

In all our working definitions, we assume that "**Insufficient Knowledge and Resources**" are the typical working conditions for any real intelligent systems, **along with the ability to adapt** (according to the definition of intelligence by Pei Wang, see above) [3, 4]. Furthermore, an advanced AGI system may additionally have a **meta-learning** (learning to learn) capability to improve gradually the learning algorithm itself, given the experience of multiple learning episodes [4, 5, 47]. It is interesting to note that, for example high AI-PQ is necessary for agile robotic and manufacturing systems, while AI-IQ intelligence is needed in all mathematically formulated problem solving systems.

## 7. AI with Social Intelligence (AI-SQ)

The main attribute, or characteristic, of **AI-SQ** are social interactions, which can be represented and realized through: ***Communication, coordination, cooperation, collaboration and co-creative collaboration*** skills – 5C skills (see for detail Figure 10).

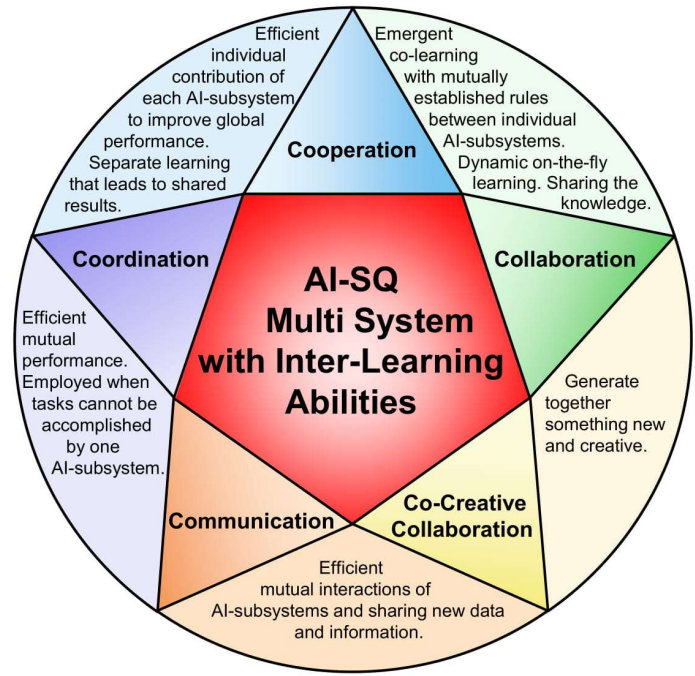

Figure 10 AI-SQ multi-system (multi-agent) with five social capabilities to interact (5C social skills): Communication, Coordination, Cooperation, Collaboration and Co-creative collaboration.

**Remark 4:** *Although words such as **coordination, cooperation and collaboration** are often used interchangeably in the context of social interactions and effective teamwork, we must note substantial differences among them. Using these words interchangeably poses a risk of confusion, as well as diluting their meaning and diminishing the potential for designing desired learning styles by AI researchers and developers* [2,11,23,35,44,45 ].

Therefore, we provide here a categorization of AI-SQ and working definitions of AI-SQ depending on their interaction levels and performed tasks:

A. ***AI with communication ability*** provides an efficient way for the exchange of information and raw data between intelligent agents.

B. By ***AI with coordination ability,*** we understand the ability of multi-agents to maintain some harmony and/or alignment among individual agents' efforts toward the accomplishment of specific common goals. Coordination can also be understood as a sequenced plan of actions to be performed by intelligent agents, by delineating who will do what, when, and within what time duration.

C. ***AI with cooperation intelligence*** is a network of multi-agents or physical smart robots, where each individual agent/robot exchanges relevant information and resources in support of each other's goals, rather than a shared common goal. It is interesting to note that in the case of cooperation, the result will be created by individual/independent agent/robot efforts, rather than through a collective team effort. In such a case, sub-tasks for each individual agent/robot are separated, but with a well-understood and defined global task for a network of multi-agents.

D. ***AI with collaboration intelligence*** is characterized by the ability of multi-agents to exchange not only information but also knowledge and to work together and/or with humans to produce or create something in support of a shared task. In general, collaboration is the action of working together with someone to produce or create something. Intelligent agents should share a common goal or principle to contribute jointly to perform a specific task.

E. ***AI with co-creative intelligence*** is a network of multi-agents, which has the ability to work together to produce something new, innovative and even unexpected, which has value and purpose and follows common sense. Such co-creative intelligence can be achieved by knowledge/expertise, experience, curiosity, exploration, flexibility, a strong motivation, prototyping, testing and exchange of ideas via feedback, adaptation, even wisdom for further improvements (see also Figure 9).

## 8. AI with Emotional and Social Intelligence

*Social intelligence* (SQ) can be considered an extension or *a superset of emotional intelligence* (EQ) since it is a much broader concept than emotional intelligence. In fact, in the psychology both intelligences are often integrated as EQ & SQ or briefly as ESI (emotional-social intelligence) [13,14,16,19,38,42].

AI with Emotional and Social Intelligence is referred to here as (**AI with EQ+SQ** with five fundamental abilities of an intelligent multi-agent: ***self-awareness, self-management, social awareness, social (interaction) skills and responsible decision making skills***. These skills would allow an AI multi-system with EQ & SQ not only to understand, but also to manage and perform self-regulation and social interactions (see for detail in Figure 11).

***Self-awareness of AI with EQ+SQ*** can be interpreted, as its ability of multi-agent to evaluate or assess its own behaviors (i.e., outputs, actions, decisions), strengths and weaknesses across various situations, i.e., assess its own performance, reliability and potential limitations in performing specific tasks in different scenarios, e.g., in the scenario when sensor data sets are corrupted by noise or outliers.

***Social-awareness* of AI** is the capacity of understanding and assessing of behaviors, or performance and/or reliability and robustness of interconnected and interacted multi-agents with human users, in respect to environmental changes, quality of input/output data, and/or the changing of desired tasks.

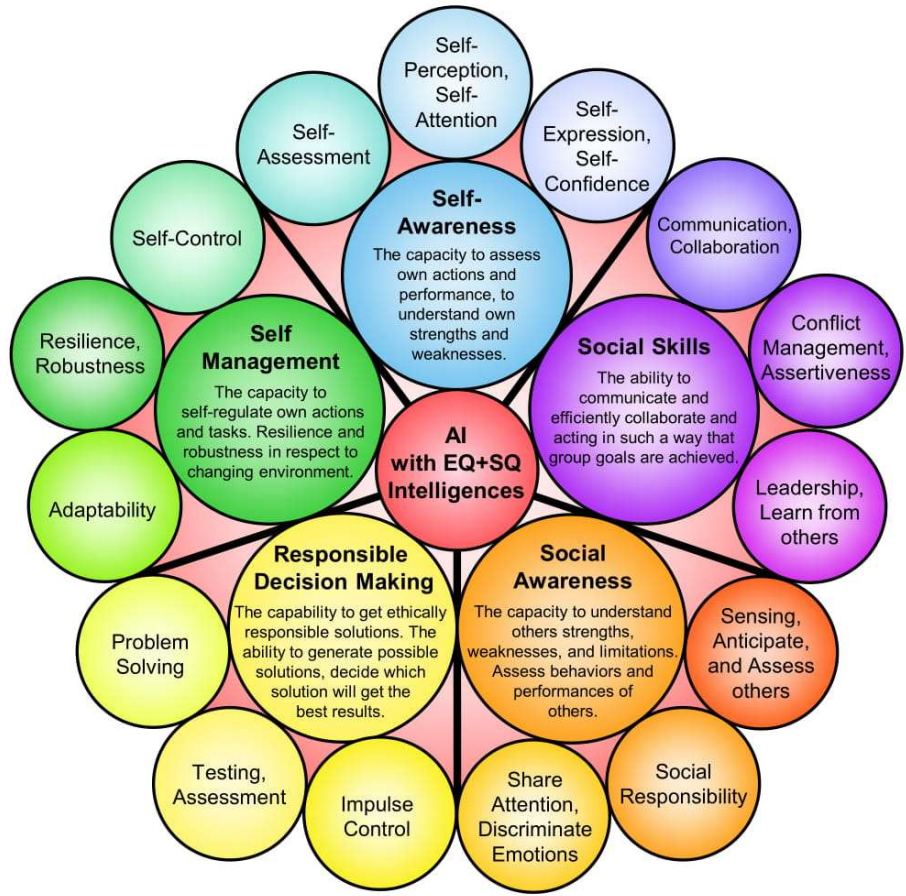

Figure 11 Conceptual graph illustrating AI with EQ+SQ or AI with ESI (Emotional-Social Intelligence) with the advanced cognitive functions: Self-Awareness, Social (external) Awareness, Interaction (Social) Skills and Self-Management abilities. Such advanced AI has both autonomous and social features.

***Self-management* of AI** is characterized as the capacity of planning and self-controlling one's own actions dependent of external changes, self-adaptability to various conditions and flexibility to a changing environment, resilience and robustness to noise and/or outliers or corrupted sensory signals/data sets and the ability to identify potential problems/conflicts in order to make a responsible choice from among the possible options.

***Responsible (ethical) decision making of AI*** is characterized by the capacity to generate alternative solutions and apply criteria which allow an intelligent agent to choose the ethical solutions or decisions, i.e., the ability to plan and act ethically and responsibly (see Figure 12). By ethical principles and solutions, we understand here: Value of humans, avoiding harm, solidarity, justice, fairness and social responsibility for common good [12].

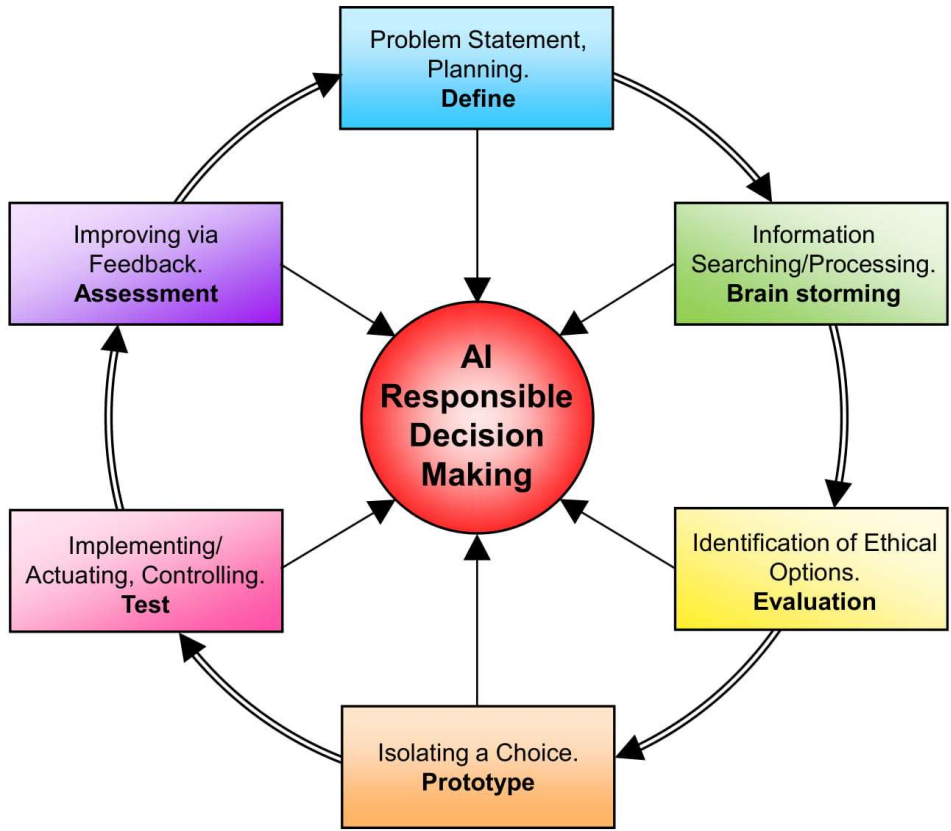

Figure 12 Responsible-ethical decision making process.

Finally, **Social interaction skills of AI with EQ+SQ** can be expressed, first all, by the abilities of communication, cooperation, collaboration, or more advanced, co-creative collaboration skills with smooth and efficient interactions and teamwork. However, other important cognitive skills could be also taken here into account when developing and implementing this type of AI, such as assertiveness, tolerance to the limitations of others, the ability to avoid or mitigate conflicted actions or solutions (i.e., efficient conflict management), and also the ability to learn from others and teach others [13-19].

## 9. Attentional Intelligence

There are several vital higher order cognitive abilities for AI, encompassing different aspects of intellectual functions and processes, including perception (visual, auditory, tactile), attention (attending specific information and ignoring other), responsible inhibition (ability to suppress inappropriate responses), inference (i.e., a conclusion or idea reached on the basis of evidence and reasoning and/or the process of reaching such a conclusion), the formation of knowledge, pattern recognition, episodic memory (association of events with place and time), short-term and long-term memory, judgment and evaluation, reasoning and computation, planning, strategic problem solving, continual meta learning, responsible decision making, comprehension and generation of language (see Figure 13 for detail).

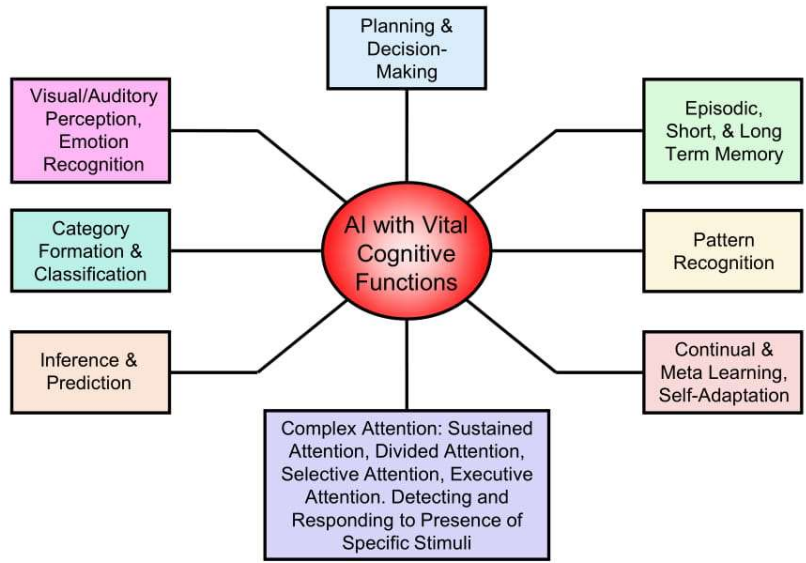

Figure 13 The eight core cognitive skills important for future AGI systems.

In this battery of important cognitive abilities and skills as regards AI, *complex attentions*, *continual meta learning and self-adaptation* to the surrounding environment are the ones which will play key roles [7, 14, 15]. **Attention** in AI can be interpreted as a *neural attention* mechanism which, for example, equips, an ensemble of deep neural networks with the ability to focus and perform a smart selection on a subset of their inputs (or features) [24-29]. For example, an AI system with attention have the ability to automatically select specific inputs or a specific subset of stimuli or input data (e.g., some specific patches of images or specific frequency of audio signals), in order to solve a problem more efficiently and/or more robustly with respect to noise or outliers.

Drawing on research in cognitive science, we can say that humans have at least four main types of attention used in our daily lives: **Selective attention, divided attention, sustained attention, and executive attention**. All these "attentions" can principally be implemented and employed in AGI systems [24-29]. Selective attention is the ability to focus or concentrate on a task even when some distractions are present (e.g., noise, outliers, changing environmental conditions) (see e.g., [28] and Figure 14 for more detail). *Alertness* is a state of being ready to react immediately to a specific stimulus, while attention mechanism in AI is principally focus on a certain specific part of information or stimuli or training data sets, when processing a large amount of raw information. *Spatial Attention* is a form of **visual attention** that involves directing attention to a location in 2D or 3D space, while **Temporal Attention** is a special case of attention (e.g., auditory attention) that involves directing attention to specific instants of time. The essence of AI with *temporal attention* is to flexibly focus in time to recognize temporal e.g., rhythmic patterns **Attention Switching** task is paradigm requiring AI system to *switch* between performing multiple different individual tasks. It can be interpreted as a perceptual cognitive function that involves the ability to unconsciously shift *attention* between one task and another. **Divided Attention** is a type of simultaneous focus that allows AI sytem to process different information sources and efficiently perform multiple tasks simultaneously, while **Executive Attention**, refers to ability to control responses, particularly in conflict situations, where several responses are possible. **Interference Suppression** is mechanism to ignore some salient perceptual information in a bivalent task while attending to the less salient conflicting information. On the other hand, **Inhibition** involves the ability to avoid further processing of stimuli or information, which could or should be ignored. ***Supervisory Attentional* Control** is a higher level cognitive mechanism active in non-routine or novel situations; it requires conscious control in response to specific environmental stimuli and uses flexible strategies to solve a variety of difficult conflicted problems. ***Meta Attention or Meta Focus*** consists of regulation of attention and knowledge of attention (i.e., noticing where AI-system focus is directed and self-awareness of employing specific strategies, so that it keeps its attention focused on the task at hand. ***Meta Memory*** is awareness of memory strategies that work best for AI system. ***Meta Perception or Meta Sensing*** means noticing what AI system is sensing/measuring or "feeling", and finally, the most sophisticated ***Meta Cognition*** involves of self-awareness of the strategy an AI system is using to learn to perform specific tasks and evaluating whether this strategy is sufficiently effective for specific tasks (see Figure 14).

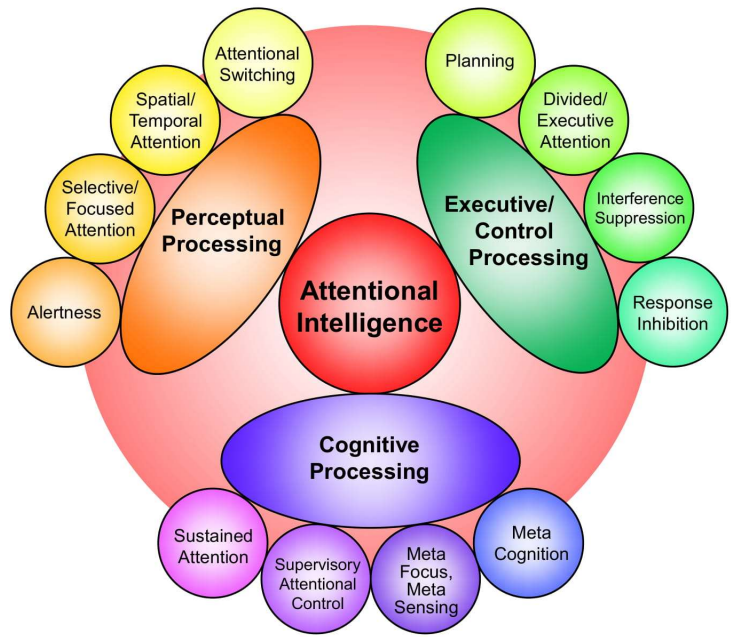

Figure 14 Components of attentional intelligence which could be key factors in future AGI systems.

Note that **Selective attention** occurs when the awareness – whether that is visual, auditory, or tactile – is channeled onto something specific or focused on relevant targets, while **Divided attention** occurs when mental focus is directed towards multiple tasks or ideas at once. On the other hand, **Sustained attention** is the ability of AI to attend to a task continuously for an extended period and **Executive attention** refers to our ability to regulate responses or decisions, particularly in situations of conflict or when AI receives confusing and contradictory stimuli. When utilizing their executive attention in such a conflict setting, a human being or an AI system should have the ability to regulate their responses accordingly, where several non–consistent responses are possible. In general, Attention can be considered as focused self–awareness and it is attracted to selected range of features of specific stimuli like images, sounds, and words. ***Attentional Intelligence (AI–AQ)*** (see Figure 14 for detail) is closely associated with the efficient processing of information and knowledge and it plays a key role in the human intelligence (cf. [23-29]).

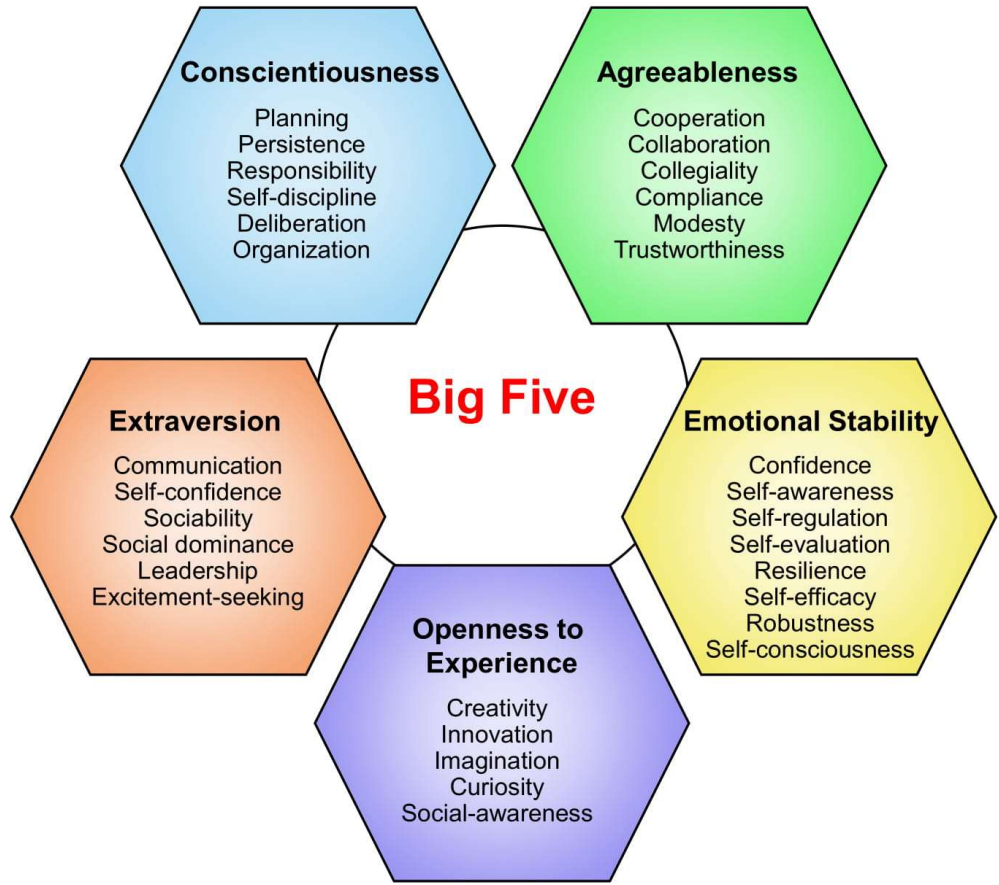

Figure 15  Links between "**Big Five**" personality traits and human multiple intelligences. The **Big Five** model is often considered to be the most robust way to describe human personality differences and it can promote the formation/discrimination of individual differences in human intelligence.

**Remark 5.** It is  interesting to note that many features,  especially self-control, self-awareness, social emotions, attention, responsible decision making, ethical-awareness, and moral–ethical responsibilities, of the proposed  AI with PQ, IQ, EQ, SQ, CQ, INQ, AQ and MQ intelligences  are  associated with the *Big Five personality traits* of humans, as also known as the five-factor model (FFM) and the OCEAN model [40-42] (see Figure 15): *Openness* (tendency to creativity, curiosity and imagination), *Conscientiousness* (tendency to being diligent, responsible and self-aware), *Extraversion* (tendency towards sociability, assertiveness and emotional expressiveness), *Agreeableness* (tendency towards being collaborative and reliable) and *Emotional Stability* (tendency towards to have robust and stable behaviors/performance and emotions, self-regulation, resilience).

## 10.  Conclusions and Discussion

There are various definitions of human and AI intelligences and creativities that have been developed and built up over years of discussion and disputing, rewording and reworking, among psychologists, philosophers, neuroscientists, cognitive and computer scientists [4,7].

Since AI research is inspired by human intelligence, we believe that multiple intelligences and corresponding learning styles will play an important role in the research, development and evaluation of a new generation of distributed AI/AGI systems with 'a human face' [48]. Furthermore, in many specific applications of AI, for example, in biomedical applications, an extremely vast diversity of knowledge and cognitive skills is required, and therefore many different forms of cognitive skills and/or intelligences could be potentially useful.

Although current state-of-the-arts AI systems already exploit and mimic some human intelligences, still, emotional, social, attentional and moral-ethical intelligences are not implemented to their full potential. For example, current AI systems have the ability, to some extent, to detect and recognize human emotions, but so far, they do not possess self-awareness, self-management, self-assessment, social-awareness and social skills to interact with other agents efficiently. Furthermore, current AI systems still have quite limited cognitive skills in other domains and are not yet able to perform intelligent and responsible decision making.

The main objective of this paper is to consider AGI systems with a more 'human face'. Since emotional-social intelligence, creative-innovative intelligence, attentional intelligence and moral-ethical intelligence related to responsible decision making are the essence of human relationships and are essential for effective teamwork and social co-existence, we expect more research and development in AI systems which will have the ability to meaningfully interact with users socially, understand their behaviors and abilities, and even understand (to some extent) theory of human minds, including complex cognitive tasks, emotions and human social interactions.

In this paper, we attempted to categorize various AI systems depending on their abilities, learning styles and learning algorithms. The essential purpose of our categorizations of AI systems (and their corresponding working definitions) is attempt to make them, as much as possible, useful, inspiring and insightful, due to the following reasons [4]:

a) They explain what kind of features or components should each specific AI system have and they have some explanatory power, which may lead to progress not just in AI but also in computational neuroscience.

b) They not only categorize AI systems, but can also allow us to measure their degree of intelligence. If there are different kinds of intelligences, we need some taxonomy for identifying the kind of intelligence possessed by a system (if any), and quantitatively comparing it to that of other AI systems.

c) They could be some kind of guide to measure the progress and/or to demonstrate some potential in the development of new generation of AI systems. Here, a key point is to measure the cognitive skills, flexibility and meta-level learning capability, rather than only the concrete problem-solving capability.

d) Furthermore, they allow us to formulate explicitly or implicitly new challenging sub-problems in AI, according to the motto of "a problem well-stated is half solved".

However, it is neither necessary nor practically useful to attempt to develop and design current practical oriented AI systems, which would be able to simulate or mimic exactly all human, multiple intelligences; this is also neither feasible nor realistic. Rather, it is desirable and expected that the next generation of AI systems would have complementary and/or augmented intelligences to existing human multiple intelligences. This concept is related to the recently introduced AI *augmented intelligence*, where AI works together with humans to enhance cognitive performance, including learning, decision making and forming new experiences. Intelligent augmentation will use and integrate human multiple intelligences, together with more advanced cognitive skills and computational technologies, but with the main objective, of not replacing humans, but rather assisting them and enhancing their capacities. For example, an AI multi-agent with emotional-social intelligence would be able to analyze social cues and human interactions so as to enhance human team collaboration. Another example could be AI agents who collaborate with human game players in E-sports to complete custom-designed missions.

## Acknowledgment

This research was supported by the Ministry of Education and Science of the Russian Federation (grant 14.756.31.0001).

**Appendix. Categorization of the State-of the Arts Machine Learning Algorithms: Supervised, Unsupervised, Reinforcement Learning, Ensemble Learning, Deep Learning and Deep Reinforcement Learning**

(a)

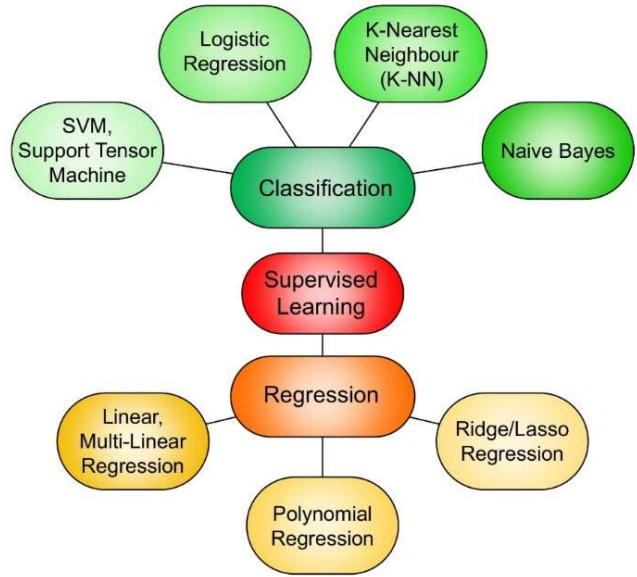

(b)

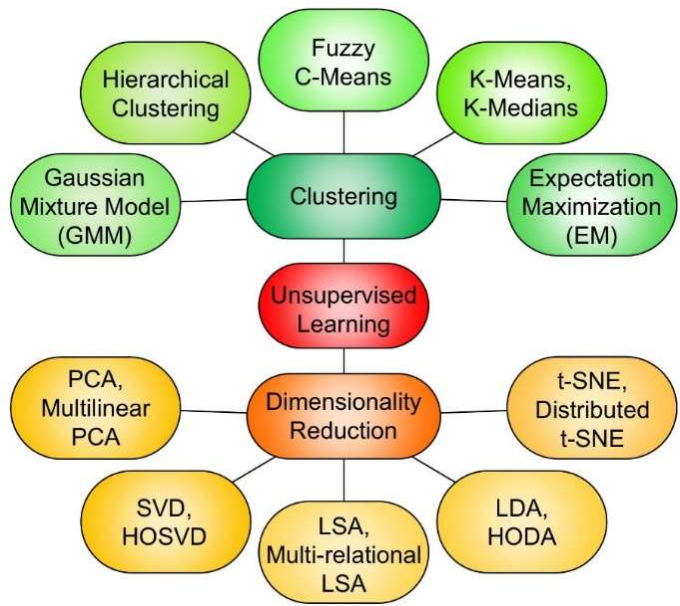

Figure 16 Basic supervised (a) and unsupervised (b) learning algorithms used in machine learning (see e.g., [5, 52]).

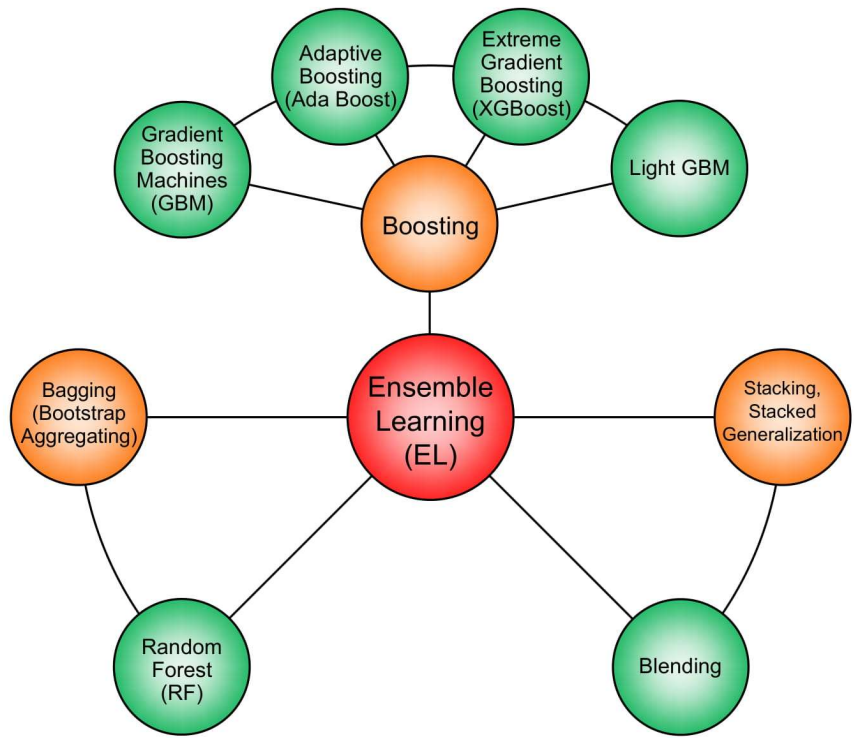

Figure 17 Basic ensemble learning algorithms.

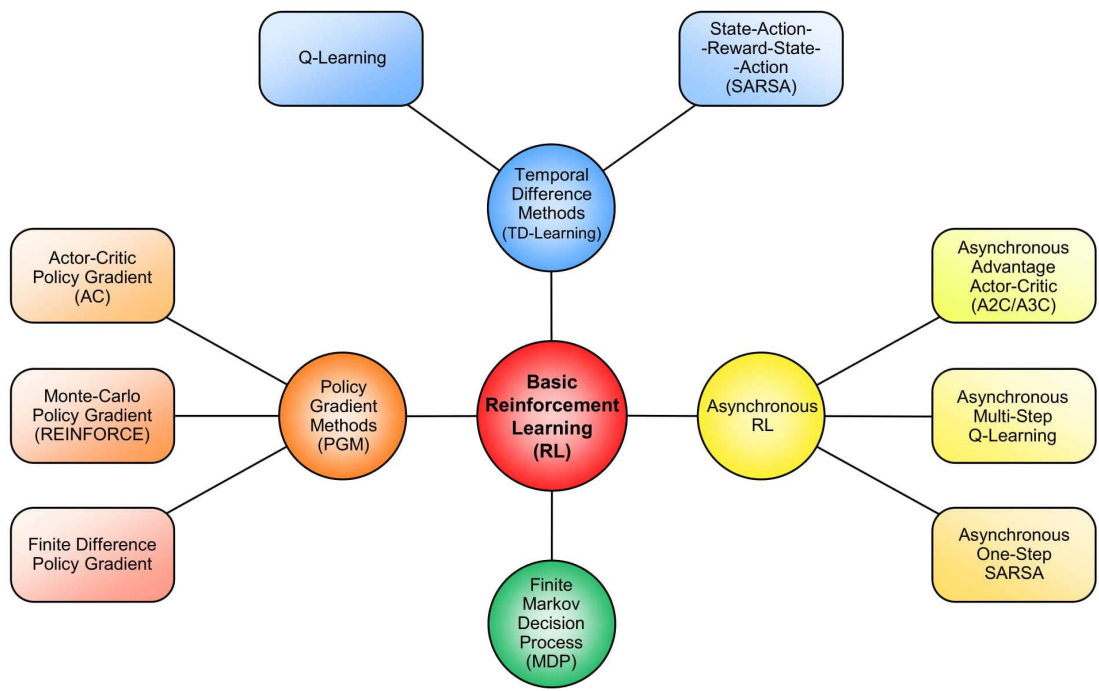

Figure 18 Basic reinforcement learning algorithms (see e.g., [20, 21]).

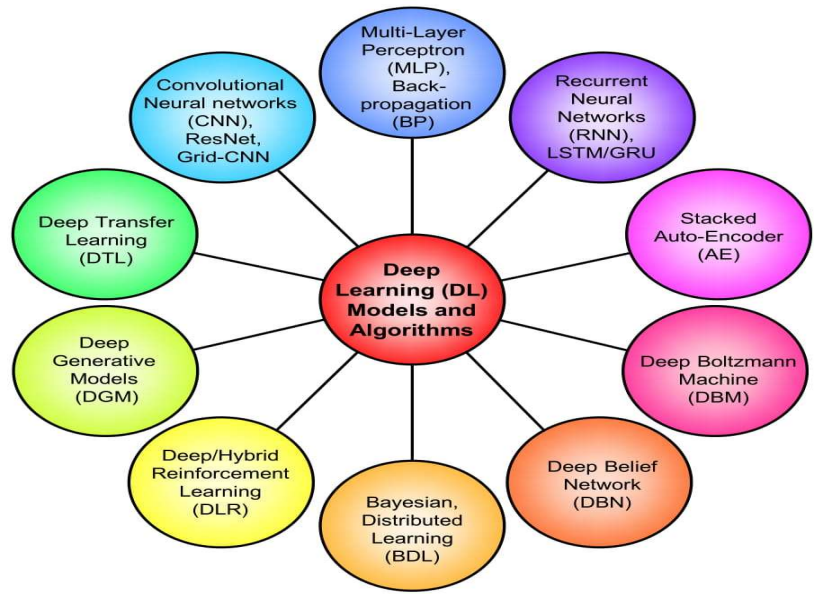

Figure 19 Basic deep leaning models and learning algorithms (meaning of abbreviations: **ResNet**- Residual Network or Residual CNN,  **LSTM** – Long Short-Term Memory and **GRU** –Gated Recurrent Unit which are special cases of RNNs).

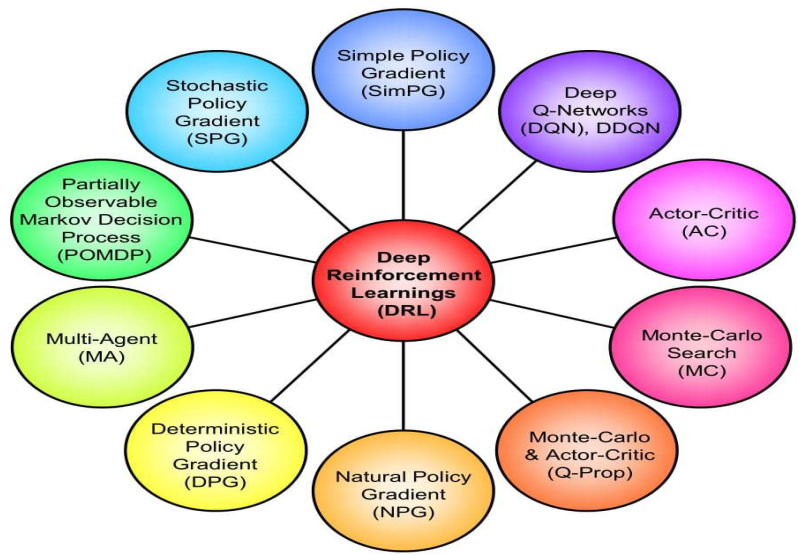

Figure 20 Categorization of deep reinforcement learning algorithms. Each category consists of a number of specific algorithms (see e.g., [20]): **Stochastic Policy Gradient** (**SPG**): REINFORCE, Soft Actor-Critic (SAC), Asynchronous Advantage Actor-Critic (A3C);  **Simple Policy Gradients algorithms** (**SmpPG**): REINFORCE, SAC, A3C, Deep Deterministic Policy Gradient (DDPG), Distributed Distributional Deep Deterministic Policy Gradients (D4PG), and Twin Delayed Deep Deterministic (TD3); **Deep Q-Networks**: DQN, Double Deep Q Network (DDQN), DDQN with Duel Architecture and DDQN with Proportional Prioritization; **Actor-Critic** (**AC**): SAC, A3C, Deep Deterministic Policy Gradient (DDPG), D4PG, TD3, Trust Region Policy Optimization (TRPO) and Proximal Policy Optimization (PPO);   **Monte Carlo** (**MC**): REINFORCE, PPO and TRPO;   **Natural Policy Gradient** (**NPG**): TRPO, PPO, Actor-Critic using Kronecker-Factored Trust Region (ACKTR) and Actor-Critic with Experience Replay (ACER);   **Deterministic Policy Gradient** (**DPG**): DDPG, D4PG and TD3;   **Q-Prop** (hybrid of MC & AC).   Finally, the most important and perspective for our AI multi-systems with multiple intelligences are two class of sophisticated RL algorithms:   **Partially Observable Markov Decision Process** (**POMDP**): Deep Belief Q-network (DBQN), Deep Recurrent Q-network (DRQN), Recurrent Deterministic Policy Gradients (RDPG), and   **Multi-Agent** (**MA**) learning: Multi-Agent Importance Sampling (MAIS), Coordinated Multi-Agent DQN, Multi-Agent Fingerprints (MAF), Counterfactual Multi-Agent Policy Gradient (COMAPG) and Multi-Agent DDPG (MADDPG) (see [20, 15, 21] ) and references therein for more details).